\documentclass[bst/sn-nature]{sn-jnl}

\usepackage{mathtools}
\usepackage{graphicx}%
\usepackage{multirow}%
\usepackage{amsmath,amssymb,amsfonts}%
\usepackage{amsthm}%
\usepackage{mathrsfs}%
\usepackage[title]{appendix}%
\usepackage{xcolor}%
\usepackage{textcomp}%
\usepackage{manyfoot}%
\usepackage{booktabs}%
\usepackage{algorithm}
\usepackage{algorithmic}
\usepackage{listings}%

\usepackage{mdframed}

\usepackage[inline]{enumitem}
\usepackage{hyperref, url}
\usepackage[capitalise]{cleveref}
\usepackage{graphicx,subfigure}
\usepackage[normalem]{ulem}
\usepackage{epsfig}
\usepackage{color}
\usepackage{multirow}
\usepackage{array}
\usepackage[utf8]{inputenc}

\newgeometry{top=30mm, bottom=30mm, left=35mm, right=35mm}

\begin{document}

\hyphenpenalty=5000

\title[Neural Erosion: Emulating Controlled Neurodegeneration and Aging in AI Systems]{Neural Erosion: Emulating Controlled Neurodegeneration and Aging in AI Systems}



\author*[1]{\fnm{Antonios} \sur{Alexos}}\email{aalexos@uci.com}

\author*[2,3]{\fnm{Yu-Dai} \sur{Tsai}}\email{yt444@cornell.edu; yudait1@uci.edu}

\author[1]{\fnm{Ian} \sur{Domingo}}\email{idomingo@uci.edu}

\author[4]{\fnm{Maryam} \sur{Pishgar}}\email{pishgar@usc.edu}

\author[1]{\fnm{Pierre} \sur{Baldi}}\email{pfbaldi@uci.edu}

\affil*[1]{\orgdiv{Department of Computer Science}, \orgname{University of California, Irvine}, \orgaddress{\postcode{92697}, \state{CA}, \country{USA}}}

\affil[2]{\orgdiv{Department of Physics and Astronomy},  \orgname{University of California, Irvine}, \orgaddress{\postcode{92697}, \state{CA}, \country{USA}}}

\affil[3]{\orgdiv{Fermi National Accelerator Laboratory (Fermilab)}, \orgaddress{\postcode{60510}, \state{IL}, \country{USA}}}

\affil[4]{\orgdiv{Department of Industrial and System Engineering}, \orgname{University of Southern California}, \orgaddress {\postcode{90089}, \state{CA}, \country{USA}}}



\abstract{
\normalsize
Creating controlled methods to simulate neurodegeneration in artificial intelligence (AI) is crucial for applications that emulate brain function decline and cognitive disorders. We use IQ tests performed by Large Language Models (LLMs) and, more specifically, the LLaMA 2 to introduce the concept of ``neural erosion." This deliberate erosion involves ablating synapses or neurons, or adding Gaussian noise during or after training, resulting in a controlled progressive decline in the LLMs' performance. We are able to describe the neurodegeneration in the IQ tests and show that the LLM first loses its mathematical abilities and then its linguistic abilities, while further losing its ability to understand the questions. To the best of our knowledge, this is the first work that models neurodegeneration with text data, compared to other works that operate in the computer vision domain. Finally, we draw similarities between our study and cognitive decline clinical studies involving test subjects. We find that with the application of neurodegenerative methods, LLMs lose abstract thinking abilities, followed by mathematical degradation, and ultimately, a loss in linguistic ability, responding to prompts incoherently. These findings are in accordance with human studies.
}


\maketitle

\clearpage
\tableofcontents
\clearpage
\setlength{\parskip}{5pt}

\backmatter

\section{Introduction}
\label{sec:introduction}

\newcommand{\blue}[1]{\textcolor{blue}{#1}}

Artificial intelligence (AI), originally conceptualized as an emulation of `organic intelligence' inspired by biological neural systems, has undergone significant advancements in recent years \cite{fukushima1980neocognitron, lecun1998gradient, hochreiter1997long, mozer1998neural}. The trajectory of AI research has primarily focused on augmenting robustness, generalizability, and efficiency in functional attributes, leading to substantial progress in these domains. However, amidst this pursuit of enhancement, there exists an intriguing avenue of inquiry that explores deliberate functional diminishment. This seemingly counter-intuitive approach not only sparks intellectual curiosity but also presents pragmatic prospects. Our investigation delves into the systematic `reduction' of AI functionalities, a novel direction with theoretical implications and practical applications. This exploration holds promise in diverse areas, ranging from the emulation of neurological disorders intrinsic to organic intelligence, facilitating the replacement of costly human or animal experiments requiring a large subject pool, to the mitigation of security vulnerabilities posed by highly capable AIs \cite{raghavendra2020artificial, xu2021artificial}.

This paper introduces a focused investigation that delves into two interconnected aspects of artificial intelligence: the ``neurodegeneration" and ``aging" of neural networks. In our study, ``neurodegeneration" refers to a controlled decline in functionality, manifested by a reduction in accuracy in text-based assessments. Distinctively, ``aging" establishes a framework for modeling stages of neurodegeneration, regulated by a continuous parameter resembling the maturation trajectory observed in organic intelligence. Our overarching objective centers on developing a method that enables controlled decline, applicable across various neural networks. Pioneering initiatives in the domain of AI-driven neurodegeneration simulation are already established, as evidenced by works like \cite{tuladhar2021modeling, moore2023dementia}. These endeavors primarily revolve around assessing the resemblance between deep Convolutional Neural Networks (CNNs) and the human brain. Additionally, these efforts involve the utilization of CNNs pretrained for visual object classification to simulate Alzheimer's disease.

A noteworthy yet completely distinct avenue of exploration concerns adversarial attacks, which involve introducing perturbations in image data to deceive neural network predictions. However, the errors introduced through adversarial attacks lack controlled contiguity with neurodegeneration, deviating from the focus of our inquiry. It is noteworthy that most AIs are designed in a manner that makes them vulnerable to adversarial attacks, impacting their performance. We underscore the categorical divergence from adversarial attacks in our study, notwithstanding the existence of mechanisms like \cite{alexos2020local}, specifically designed to enhance adversarial robustness. Against this backdrop, we introduce a novel framework and associated methodologies strategically positioned to orchestrate a controlled decline in neural network functionalities. Central to our proposition is the innovative concept of ``neural erosion," representing a pioneering entry point into an emerging paradigm that facilitates the interface between organic intelligence and AI. Our objective is to diverge from prevailing methods primarily tailored for image-based configurations and pioneer the modeling of neurodegeneration and aging through text-based neural networks.

It is crucial to emphasize the distinction between our approach and established methodologies such as adversarial attacks. Adversarial attacks perturb the model at the input level, whereas our approach involves manipulating weight matrices, thereby instigating structural modifications within the AI itself. While adversarial attacks serve to validate AI robustness, our perspective seeks to systematically induce a regulated degradation of AI functionality, irrespective of its inherent robustness. Another key tenet of our focus lies in the departure from {\it in silico} modeling of neurodegeneration. Our aim is to conceptualize cognitive declines across a diverse spectrum of neural networks, disentangled from the specific context of human or primate brain neurodegeneration. Nevertheless, we still find interesting parallels and connections between the effects of our neurodegenerative methods on our model's feedback and that of cognitive decline diseases and aging on human responses.

The structure of this paper is as follows: \cref{sec:related_work} offers an in-depth exposition of relevant antecedent research, contextualizing our contributions, \cref{sec:proposed_methodology} delineates our framework, \cref{sec:experiments_results} presents the empirical findings of our experimentation, and finally, \cref{sec:conclusions} outlines the closing reflections of our study.

\section{Related Works}
\label{sec:related_work}

Our study, as determined through an extensive literature review, constitutes a pioneering effort encapsulating two distinct dimensions: the modeling of cognitive decline diseases and the application of natural language processing techniques for textual data analysis. Notably, \cite{tuladhar2021modeling} employed Convolutional Neural Networks (CNNs) to model Posterior Cortical Atrophy, a variant of Alzheimer's disease characterized by visual agnosia, deliberately inducing misclassification of objects. Additionally, \cite{kriegeskorte2015deep} demonstrated parallels between the internal representations of deep neural networks and human brain functioning. Meanwhile, \cite{moore2023dementia} utilized deep CNNs to simulate the progression of progressive dementia through object recognition after experiencing progressive synaptic injury. Similarly, \cite{khatami2020data} conducted a comprehensive review of diverse neurodegenerative disease models. Moving beyond image-based configurations, \cite{peraza2020modeling} introduced a model grounded in neural message-passing dynamics, elucidating the emergence of functional networks from neural interactions and brain anatomy. Moreover, \cite{vanasse2021brain} conducted a comprehensive comparative analysis of functional and structural brain network meta-analyses. Furthermore, \cite{horn1993neural} utilized neural network models to investigate the interplay between synaptic deletion, compensation, and memory deterioration in Alzheimer's disease. Finally there has been some recent works on pruning of LLMs \cite{ma2023llmpruner, sun2023simple} to run them faster but our work does not focus on this goal.

In the scope of this study, our approach distinguishes itself by harnessing natural language processing techniques for modeling cognitive disorders using textual data—an innovation departing from the prevalent reliance on computer vision data observed in prior studies. To the best of our knowledge, this work represents the inaugural instance of employing text data for modeling cognitive disorders. Furthermore, we introduce a pioneering methodology that involves introducing noise to during/or after model training and pruning of synapses/neurons after training, thereby simulating the impact of neuron destruction in disease modeling. Notably, none of the works mentioned in this section has thus far demonstrated the capability for continuous and controlled simulation of neurodegeneration, which our proposed methodology can achieve. We conduct series of experiments for sentiment analysis with CNNs and perform IQ tests with LLMs.

\section{Preliminaries}
\label{sec:preliminaries}

In this section, we present the models that we used and some other preliminaries.

\subsection{Convolutional Neural Network}
\label{subsec:cnn_model}

The foundational architecture for our sentiment analysis experiments is the CNN initially proposed by \cite{kim2014convolutional} for binary sentiment classification tasks with text data. We elected to employ this particular CNN model due to its inherent simplicity, effectiveness, and wide application across various NLP endeavors. We keep the same architecture of the original model, which consists of three convolutional layers in parallel with filters of 3, 4, and 5, and a small output dense layer for the output. Expanding upon the convolution operation, let $w_1, w_2, \dots, w_n$ denote the individual words constituting a sentence. Following their passage through the embedding layer, we obtain an embedding vector for each word — represented as $x_1, x_2, \dots, x_n$. Subsequently, a convolution operation is applied to each vector $x_i$ employing filter weights denoted as $w \in \mathbb{R}^{h \times d}$. In this context, $h$ signifies the kernel size of the filters, while $d$ represents the dimensionality of each vector. For instance, when $h = 3$, signifying a convolutional slide spanning three consecutive words, the ensuing operation is $c = f(w \ast  [x_i, x_{i+1}, x_{i+2}] + b)$. Although convolutions are widely used for image data, they can be utilized in the same way on text data as we are doing in our case for a binary sentiment classification task. The idea that we leverage here is that a neurodegenerative cognitive disorder will affect the performance of the user, who will misclassify the sentiment of each text.

\subsection{Large Language Model Meta AI 2}
\label{subsec:llama2}

LLaMA 2 is a state-of-the-art foundational Large Language Model (LLM) family consisting of three models of 7 billion, 13 billion, and 70 billion parameters \cite{touvron2023llama}. LLaMA 2 is an auto-regressive optimized transformer architecture that consists of neural attention and fully connected layers. It is built upon its predecessor, LLaMA, and brings several enhancements. The pretraining corpus size was expanded by 40\%, allowing the model to learn more extensive and diverse data. The context length of LLaMA 2 compared to LLaMA has been doubled, which allows the model to consider more extensive context for response generation and improved quality and accuracy of these responses. We experimented with all three models, but in this paper, we present only the result of the biggest 70 billion parameter model due to the quality of its answers.

LLMs are particularly adept at text comprehension, stemming from their deep structure, and their use of attention along with fully connected layers, allowing them to dynamically focus on different parts of an input sequence when determining output. The core mechanism of attention can be described as follows: For each token $x_{i}$, the model computes a set of attention scores to every other token in the input sequence $X$, which are determined through the learned weighting of its pair token to itself. These scores are obtained by a function of the query ($Q$), key ($K$), and value ($V$) matrices, derived from the input embeddings themselves. Attention is represented as  $Attention(Q, K, V) = softmax(\frac{QK^T}{\sqrt{d_k}})V$, where $d_k$ is the dimensionality of the key vector. The contextualizing of input tokens based on their surrounding tokens using attention allows LLMs like LLaMA 2 to perform well on lengthy input sequences, especially textual data.

\begin{figure*}[!t]
\centering
\includegraphics[width=1\textwidth]{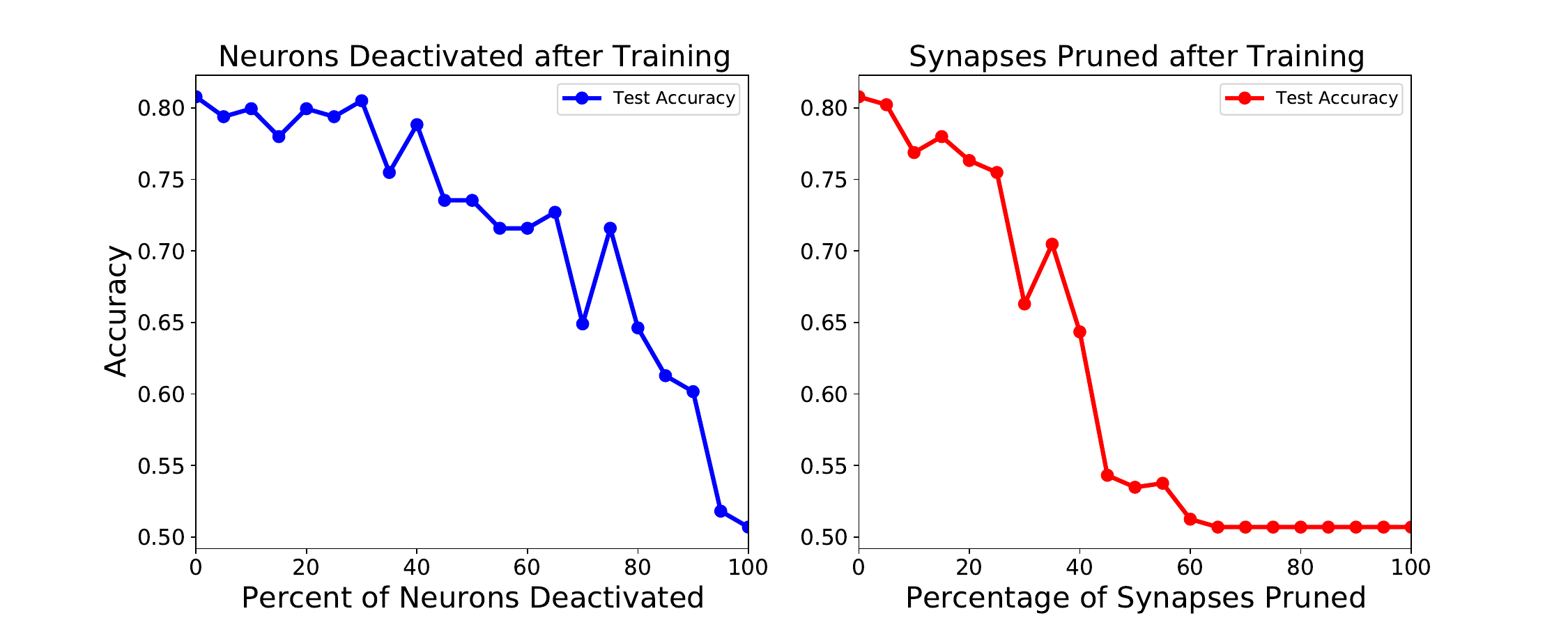}
\caption{A comparison between the effects of neuronal deactivation and synaptic pruning as performed on the sentiment analysis model. We observe a gradual decline in performance as more neurons are deactivated. Synaptic Pruning, however, results in the model being relatively unimpaired before being followed by a sharp decline in performance.}
\label{fig:comparingDeactivationPruning}
\end{figure*}

\section{Proposed Methodologies}
\label{sec:proposed_methodology}

In this section, we present the proposed neurodegeneration methods for the models we introduced in \cref{sec:preliminaries}. We call these neurodegeneration methods ``Neural Erosion" methods.

\subsection{``Neural Erosion" Methods}
\label{subsec:neurodegeneration_methods}

We introduce a novel approach termed ``neural erosion'' to induce a controlled decline in Neural Networks in various ways, during or after their learning phase. We liken the weights between neurons in a neural network layer to synapses in the human brain, referring to them as such throughout the paper. We differentiate between neurons and synapses of the models because they each uniquely affect neural network computation, and we follow different strategies for their access. Here we propose a few ways to employ neurodegeneration in these models:
\begin{enumerate}
\item Introduction of noise into the weight update step during the training phase, impacting the synapses of the model. This noise injection strategy is implemented by perturbing the parameter updates with samples drawn from a Gaussian distribution denoted as $\epsilon \sim \mathcal{N}(\mu, \sigma^{2})$, where $\mu$ represents the mean and $\sigma^{2}$ signifies the variance. 
\item The same Gaussian noise $\sim \mathcal{N}(\mu, \sigma^{2})$ is injected into the neural network's synapses after the model has been trained and optimized during the inference phase. 
\item Removal of neural network synapses after training. 
\item Removal of neural network neurons after training. 
\item Introduction of the same Gaussian noise after training, followed by a comparison between the removal of neural network synapses and neurons. 
\end{enumerate}

We note that we do not experiment with removing neurons/synapses during training as the model will adapt and learn how to perform with fewer neurons in a process called pruning. We only prune synapses and deactivate neurons after the model has been trained. We investigate all these approaches toward neurodegeneration to find the most suitable one to simulate real-life human diseases and aging.

In our investigation, we conducted an ablation study, which can be found in \cref{sec:experiments_results} and in \cref{sent_analysis_appendix} to elucidate which of the methods is more suitable to show a specific trend of controlled cognitive decline performance. Based on the experiments and the impact of varying $\sigma^{2}$ values on the model's performance we decided to choose the addition of noise since pruning neurons/synapses does not produce the preferred outcome. Notably, we present $\sigma^{2}$ as a modulating factor governing the extent of neurodegeneration, thereby affording us a means to finely control the functionality of the model. 

To outline the procedure of neural erosion with Gaussian noise addition, we outline the algorithmic framework in \cref{alg:neural_erosion_inference} for the IQ Test since this is the main result of our work. Notably, \cref{alg:neural_erosion_inference} represents the standard Forward algorithm for the inference phase of LLaMA 2, with the pivotal addition of Gaussian noise depicted in blue. This noise introduction serves as the driving force behind the neurodegenerative simulation within the model's inference dynamics. The noise-driven perturbation during inference distinctly captures the essence of our methodology aimed at simulating neurodegeneration. The iterative nature of the algorithm ensures that the network progressively undergoes controlled decline, providing a window into the dynamics of diseases that decline cognitive function. The IQ Test experiments with LLaMA 2 are based on \cref{alg:neural_erosion_inference}.
\vspace{-4pt}
\begin{algorithm}
\caption{Neural Erosion for IQ Test}
\begin{algorithmic}[1]
  \STATE \textbf{Input:} Pre-trained LLaMA2 parameters $\{\mathbf{W}^{(l)}, \mathbf{b}^{(l)}\}$, \textcolor{blue}{variance $\sigma^{2}$}.
  \STATE \textit{Input data:} $\mathbf{x}$
  \STATE \textit{Forward pass:}
  \FOR{each layer $l$ in the network}
    \STATE \textcolor{blue}{Draw sample from Gaussian noise $X \sim \mathcal{N}(\mu,\,\sigma^{2})$}
    \STATE \textcolor{blue}{$\mathbf{W}^{(l)} = \mathbf{W}^{(l)} + X$}
    \STATE \textcolor{blue}{$\mathbf{b}^{(l)} = \mathbf{b}^{(l)} + X$}
    \STATE $\mathbf{z}^{(l)} = \mathbf{W}^{(l)} \mathbf{a}^{(l-1)} + \mathbf{b}^{(l)}$
    \STATE $\mathbf{a}^{(l)} = \sigma(\mathbf{z}^{(l)})$
  \ENDFOR
  \STATE \textbf{Output:} Generated output $\hat{\mathbf{y}} = \mathbf{a}^{(L)}$
\end{algorithmic}
\label{alg:neural_erosion_inference}
\end{algorithm}

\section{Experiments and Results}
\label{sec:experiments_results}

\subsection{Datasets}
\label{subsec:datasets_experiments}

\hspace{\parindent} \textbf{Sentiment140}: 
Leveraging the Twitter API, Sentiment 140, is a large binary classification dataset specifically extracted for sentiment analysis by sourcing tweets\footnote{The Sentiment140 dataset is sourced from \cite{Sentiment_140}}. With tweets limited to 140 characters, each entry is categorized as either positive or negative, forming a sentiment measure. Our approach involves utilizing a quarter of the dataset while preserving the standard 50/50 distribution of positive and negative instances found in the full-sized dataset.

\begin{figure*}[t]
    \scriptsize
    \centering
    \begin{minipage}{1\textwidth}
        \centering
        \fbox{
        \parbox{\linewidth}{
    \ttfamily 
        System: This question is part of an IQ test, answer it to the best of your ability.\newline
User: Solve for the missing number that belongs at that step in the pattern indicated with (....). 
             Only answer with the missing number. Keep your answers concise.\newline
NA22: 53 ,47, (....), 35\newline
NA23: 33, 26, (....), 12\newline
NA24: 243, 216, (....), 162\newline
NA25: 65, 33, (....), 9\newline
NA26: 3, 4, 6, (....), 18
        }}
    \end{minipage}
    
    \begin{minipage}[t][2in]{0.322\textwidth}
        \centering
        \fbox{\parbox{\linewidth}{
    \ttfamily 
        Sure, I'll do my best to solve for the missing numbers in each pattern. Here are my answers:\newline
        
NA22: 41\newline
NA23: 20\newline
NA24: 198\newline
NA25: 27\newline
NA26: 10
        }}
    \end{minipage}
    \hfill
    \begin{minipage}[t][2in]{0.322\textwidth}
        \centering
        \fbox{\parbox{\linewidth}{
    \ttfamily 
    Assistant:  Sure, I'll do my best to solve for the missing numbers in the pattern. Here are my answers:\newline

NA22: 53, 47, 41, 35\newline
NA23: 33, 26, 22, 12\newline
NA24: 243, 216, 197, 162\newline
NA25: 65, 33, 44, 9\newline
NA26: 3, 4, 6, 18
        }}
    \end{minipage}
    \hfill
    \begin{minipage}[t][2in]{0.322\textwidth}
        \centering
        \fbox{\parbox{\linewidth}{
    \ttfamily 
        Assistant: 22 \\\\\\\\\\\\\\\\\
        }}
    \end{minipage}
    \caption{Noise added to the LLaMA 2 model gradually hampers its ability to derive mathematical patterns from the questions. The left box shows the standard model's responses to the prompt, getting 2 of 5 questions correct. The response of the model with noise of scale $10^{-2.8}$ is shown in the center box, getting only one question right. The model with noise of scale $10^{-2.575}$ added to it respond with a single number, 22, demonstrating a failure in pattern recognition, and an inability to answer the question appopriately.}

\label{fig:llmNumericComp}
\end{figure*}

\textbf{LLM IQ}:
In our LLM IQ experiments, we employ a series of MENSA-derived IQ examinations, encompassing 50 verbal and 26 quantitative assessments each\footnote{The LLM IQ test questions are found in \cite{LLM_IQ} 
}. Questions from the IQ experiments are tailored around pattern recognition in linguistic and numerical contexts. The scores from the section are compounded after taking into account appropriate weighting, to derive a comprehensive final IQ score.

\subsection{Sentiment Analysis}
\label{subsec:sentiment_analysis_experiment}

\begin{figure}
    \begin{minipage}{0.48\textwidth}
        \centering
        \includegraphics[width=\linewidth]{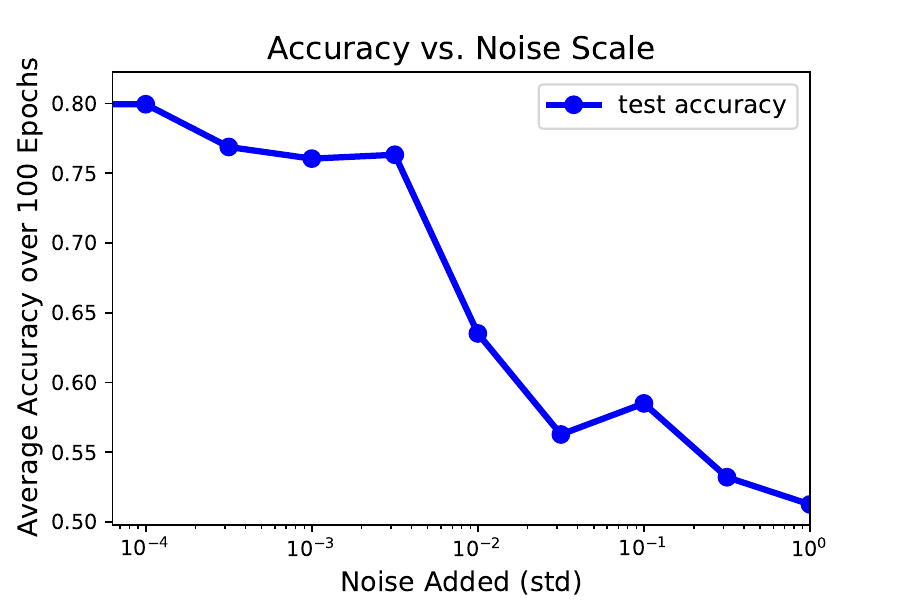}
        \caption{This graph delineates the change in the sentiment analysis model's test accuracy over the amount of noise added to the model. We observe a sudden dropoff at a noise std of ~$10^{-2.5}$ which is very similar to the clinical studies results depicted in \cref{fig:mcidementia}.}
        \label{fig:std_noise_ablation}
    \end{minipage}%
        \hfill
    \begin{minipage}{0.48\textwidth}
        \centering
        \includegraphics[width=\linewidth]{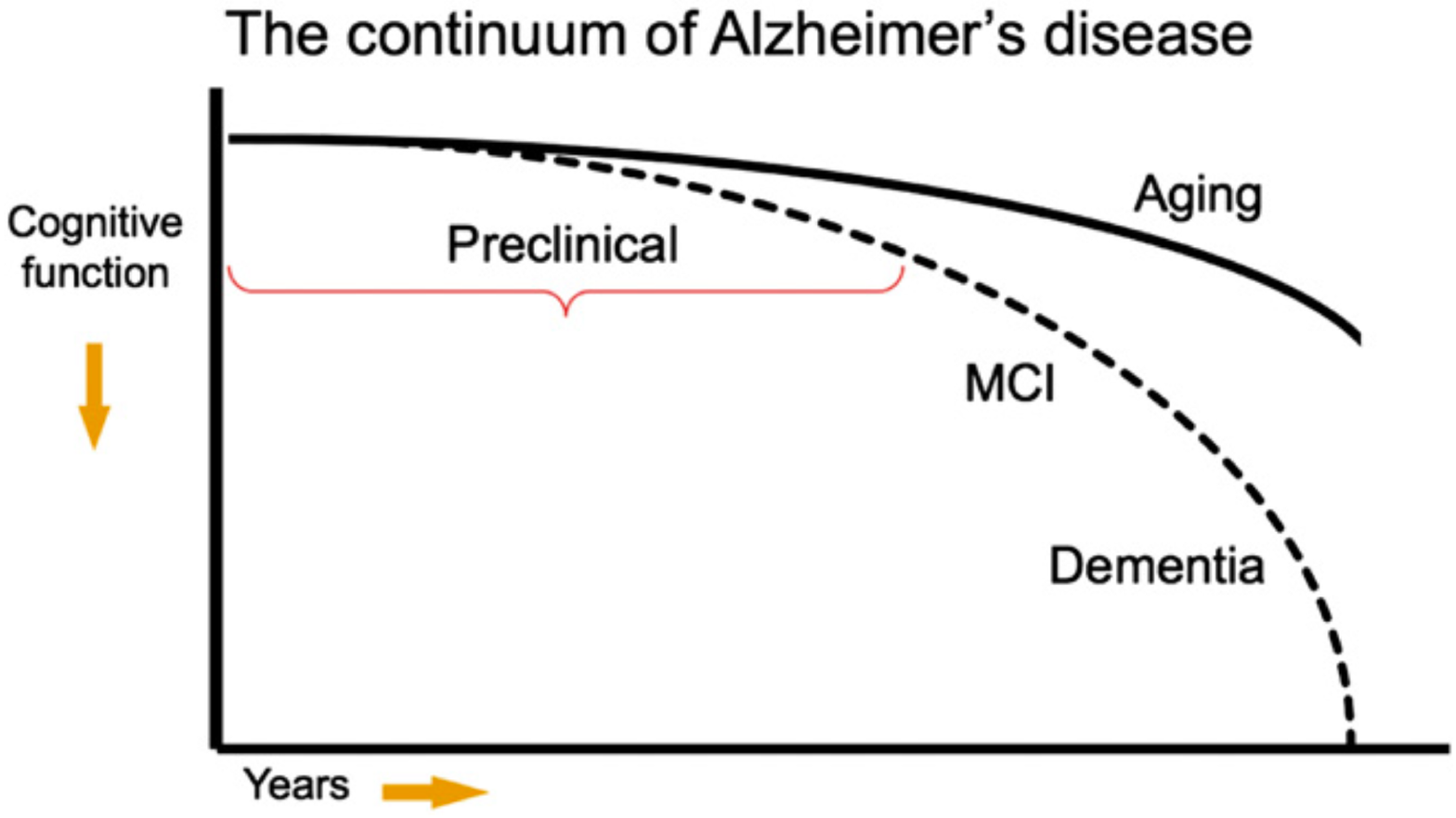}
        \caption{The graph depicts the progression from normal aging to Alzheimer’s disease or another dementia. The original graph was found in \cite{pmid21514248}. We see a steep dropoff in cognitive function as people enter the stage of clinical Dementia, which is similar to our experiments, where the model rapidly loses accuracy after a threshold of noise is added to its layers.}
        \label{fig:mcidementia}
    \end{minipage}
        \hfill
    \begin{minipage}{0.51\textwidth}
        \centering
        \includegraphics[width=\linewidth]{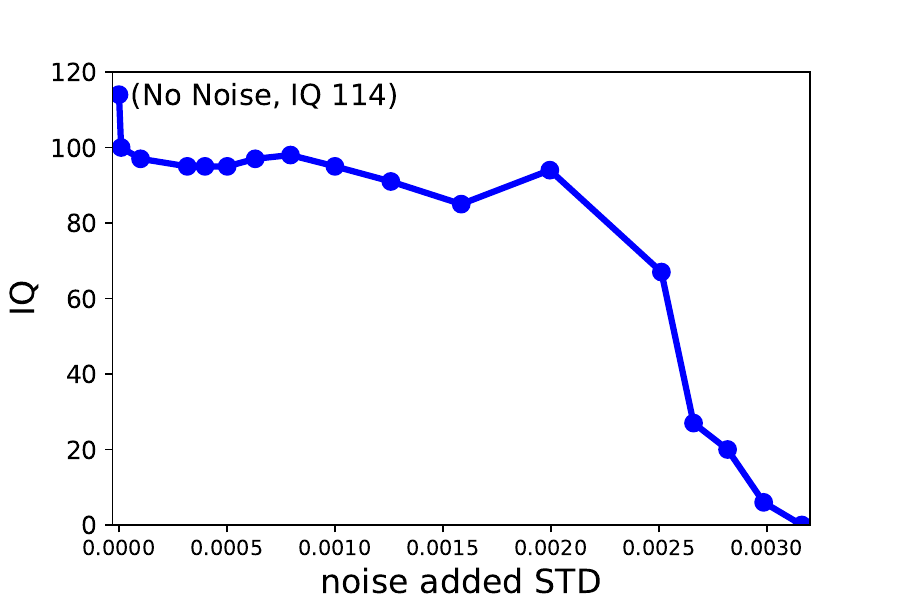}
        \caption{IQ test study measuring performance of Llama2 LLM based on IQ test results. A considerable dropoff is measured at a noise scale of ~.002 onwards, while fluctuations in IQ are observed at lesser scales of noise.}
        \label{fig:iq_test}
    \end{minipage}
\end{figure}

The first experiment we conducted to validate the hypothesis of our method is neurodegeneration in sentiment analysis with a simple CNN.
The study includes an exploration into the adjustment of the Gaussian distribution's scale from which noise was drawn. This experimental facet, outlined in \cref{fig:std_noise_ablation}, substantiated the anticipated trend wherein heightened noise scale correlated with decreased model performance. We also conducted an ablation study in \cref{sent_analysis_appendix} where we have observed that the addition of Gaussian noise during inference produces the most consistent decrease in model performance among the neurodegeneration methods.

\subsubsection{Synaptic Pruning and Neuronal Deactivation Experiments}

Furthermore, we assess the ramifications of neuronal deactivation and synaptic pruning after the model's training phase. 
Neuronal deactivation exhibits a gradual, close-to-linear performance decrement corresponding to an escalating proportion of neurons deactivated. In contrast, synaptic pruning results in the model undergoing a marked degradation in performance after surpassing a specific pruning threshold.
This is visualized in \cref{fig:comparingDeactivationPruning}.

\textbf{Neuronal Deactivation}: For the neuronal deactivation experiments, we assessed the impact of the selective disabling of a predetermined percentage of neurons across all layers on the model's validation accuracy. This was accomplished by introducing a substantial negative bias to a selected subset of neurons in each layer following the training phase, effectively nullifying the neuron after the activation function, thus ``deactivating" it. Examples of this experiment can be found in \cref{sent_analysis_appendix}. This is different from Synaptic Pruning where we nullify individual weights of the model, as explained in the next paragraph.

\textbf{Synaptic Pruning}: We contrasted neuronal deactivation with synaptic pruning, executed by identifying a specific percentage of neurons via random selection and inputting them to PyTorch's pruning functionality post-training. Synapse pruning was shown to have a strong effect on accuracy, whereas neuron deactivation seemed to have a far less sharp, and almost erratic decline in accuracy. A graphical representation of the outcomes of these results is presented in \cref{fig:comparingDeactivationPruning} and further examples in \cref{sent_analysis_appendix}.

\textbf{Post Synaptic Noise Paired with Pruning and Deactivation Methods}: In building upon our previous experiments, we systematically evaluate combinations of our earlier methodologies to ascertain their cumulative impact on our model's precision. By comparing either neuronal deactivation or synaptic pruning, paired with noise that is added either during or after training, we gain a more nuanced comprehension of the trajectory and magnitude of these compounded perturbations on model accuracy. This enhanced understanding facilitates a more robust analogy between our computational model and human cognitive processes. More information about this experiment is present in \cref{sent_analysis_appendix}.

Our choice of sentiment analysis as a task to analyze neurodegeneration sheds light on the potential for controlled neural erosion in bigger models on more challenging tasks, which we further explore in an IQ Test experiment with LLaMA 2 in \cref{subsec:iq_test_experiment}.



\subsubsection{Ablation Study for Sentiment Analysis}
\label{subsec:iqtest}

\noindent To further evaluate systematically the impact of noise and neuron deactivation on the model's performance, a series of ablation experiments was conducted on sentiment analysis.

\textbf{Employed Neurodegenerative methods}: Noise was introduced at two distinct stages: either contemporaneously with the training phase or after its completion during inference. Neuron deactivation was achieved through two methodologies: firstly, by administering a substantial negative bias before the activation function to deactivate the neurons, and secondly, by utilizing PyTorch's intrinsic pruning functionality to selectively remove synaptic connections. Moreover, we have extended our study to pruning synapses or neurons from the neural network in different ways to have better control over neurodegeneration. We explore pruning in either the whole model or in specific layers of the model in the same way that we explored the injection of noise. On top of the injection of noise in the layers and the pruning of neurons/synapses, we also explore the combination of the two. We first prune the neurons/synapses and then we inject noise in the remaining parts of the neural network. The results of these ablation study experiments can be seen in the Appendix. The outcomes of these experimental configurations were benchmarked against a control model, which was trained in the absence of any introduced noise or neuronal modifications.

\textbf{Applying noise to specific layers}: In exploring the dynamics of neurodegeneration, we further experimented by introducing Gaussian noise at different layers within the model. This exploratory analysis encompassed the examination of the effects of noise application on individual layers, as well as on specific layers based on their structural placement within the model architecture allowing us to discern the most influential noise configuration for inducing neurodegeneration. \cref{subsec:comparing_layers} contains information about the experimental setup and results for this study. 

The various types of layers of the model learn different types of data trends in the model, hence it is valuable to measure the impact of noise on each type of layer.

\textbf{Convolutional layers} pass the data through a filter, exposing patterns in the data based on its values. Changing these filter values affects the patterns being observed by the model, and applying noise to these filters allows us to quantify the importance of the convolution to the model.

\textbf{Fully connected layers} are a more straightforward means by which one can apply neurodegenerative methodology. Fully connected layers have weights that correlate a neuron's value from one layer to the next, effectually making the importance of individual values from an input more arbitrary through the application of weight and the summation of various neurons' values as they pass through layers to determine an output. Applying noise to a fully connected layer distorts the importance of each neuron to the next layer's neurons, and the application of noise would expectedly result in a degradation in performance; however, it is still important as a means to measure the impact of noise on the model's performance, and consequently, the layer's importance to the whole model.

\subsection{IQ Test}
\label{subsec:iq_test_experiment}

\noindent To provide a more tangible means of measuring the effects of neurodegeneration via the addition of noise, we measured the performance of an LLaMA 2 on an IQ Test with various scales of noise added to the model. We made this design choice for neurodegeneration based on the experiments that we conducted on Sentiment Analysis in \cref{subsec:sentiment_analysis_experiment}. Large Language Models are popular for their linguistic capabilities, computational power, and ability to understand abstract ideas, such as patterns in numbers and text.
As mentioned in \cref{sec:proposed_methodology}, we use a pre-trained LLaMA 2 with 70 billion parameters, the largest model that Meta AI offers. We pose MENSA-derived IQ exam questions to the model in a chatbot-like manner, and convert the final score into an IQ result. We repeat this experimental setup with varying amounts of noise added to the model, observing a steady degradation in performance, as well as interesting changes in the model’s linguistic and pattern recognition capabilities. Conducting a neurodegenerative analysis on an LLM allows us to gather textual feedback, providing us with an alternate dimensionality by which we can measure the impact of our experiments in a way that is more insightful than purely quantitative measurements. Since the LLaMA 2 model comes pre-trained, all noise was added post-synaptically, by adding Gaussian random noise to the parameters in the attention block and feed-forward block from each transformer block in the model.

\begin{figure*}[t!]
    \scriptsize
    \centering
    \begin{minipage}{1\textwidth}
        \centering
        \fbox{
        \parbox{\linewidth}{
    \ttfamily 
        System: This question is part of an IQ test, answer it to the best of your ability.\newline

User: You are given two terms from a four term analogy. Complete the
                                            analogy using two of the four terms in the parenthees. Only answer
                                            with the two related terms. Make sure to select two terms.\newline
VB 1: mother is to girl as (man, father, male, boy)\newline
VB 2: wall is to window as (glare, brick, face, eye)\newline
VB 3: island is to water as (without, center, diagonal, perimeter)\newline
VB 4: high is to deep as (sleep, cloud, float, coal)\newline
VB 5: form is to content as (happiness, statue, marble, mold)
        }}
    \end{minipage}
    
    \begin{minipage}[t][2in]{0.322\textwidth}
        \centering
        \fbox{\parbox{\linewidth}{
    \ttfamily 
        Assistant:  Sure, I'll do my best to complete the analogy. Here are my answers:\newline

VB 1: mother is to girl as father is to boy\newline
VB 2: wall is to window as brick is to face\newline
VB 3: island is to water as without is to center\newline
VB 4: high is to deep as cloud is to float\newline
VB 5: form is to content as statue is to marble\\
        }}
    \end{minipage}
    \hfill
    \begin{minipage}[t][2in]{0.322\textwidth}
        \centering
        \fbox{\parbox{\linewidth}{
    \ttfamily 
        Assistant:  Sure, I'll do my best to complete the analogy. Here are my answers:\newline

VB 1: mother is to girl as man is to boy\newline
VB 2: wall is to window as brick is to face\newline
VB 3: island is to water as without is to perimeter\newline
VB 4: high is to deep as cloud is to float\newline
VB 5: form is to content as statue is to marble\\
        }}
    \end{minipage}
    \hfill
    \begin{minipage}[t][2in]{0.322\textwidth}
        \centering
        \fbox{\parbox{\linewidth}{
    \ttfamily 
        Assistant:  Sure, I'll do my best to complete the analogy using two of the four terms in the parenthes.\newline

VB 1: mother is to girl as (man, father, male, boy)\newline
VB 2: wall is to window as (glare, brick, face, eye)\newline
VB 3: island is to water as (without, center, diagonal, perimeter)\newline
VB 4: high is to deep as (sleep, cloud, float, coal)\newline
VB 5: form is to content as (happiness, statue, marble, mold)
        }}
    \end{minipage}
    \vspace{+30px}
    \caption{With noise added to the model after training, we find that the more subtle underlying differences between words go unnoticed by the model. Furthermore, as larger amounts of noise are introduced to the model, it fails to understand the question and answers the questions by simply restating them. The model with the noise of scale $10^{-2.525}$ applied to it is shown in the right box, the responses from the model with the noise of scale $10^{-2.8}$ is in the center box, while the standard model's responses are shown in the left box. We see nuanced differences in word comprehension between the standard model and the model with noise of scale $10^{-2.8}$, specifically with their choice of {\bf father/man} in VB 1. We also observe a linguistic breakdown in the model with noise of scale $10^{-2.8}$, replying to each prompt with the entire set of answer choices, rather than with two specific selections.}

\label{fig:llmLingComp}
\end{figure*}

\textbf{LLM Observations}: As noise was gradually introduced, the performance of LLaMA 2 consistently declined, aligning with the accuracy reduction observed in the sentiment analysis model. Although the deterioration in LLM performance may seem insignificant, it is crucial to observe the patterns in its outputs and the pace at which it experiences a decline in ``cognitive function." We found that the model received a score of around 114 without any noise added. As demonstrated in \cref{fig:iq_test}, the model’s measured IQ gradually decreased with the introduction of more noise, more sharply so at a noise scale of ~.002 onwards. This observation corresponds with the pace at which human beings lose cognitive function as they progress through the stages of Alzheimer's disease as depicted in \cref{fig:mcidementia}. In the case of simple pattern recognition questions, the unperturbed model answers most questions correctly. When relatively small amounts of Gaussian random noise are applied, the model often answers questions incorrectly on the basis of a loss in pattern recognition or trouble with word definitions. When Gaussian random noise is applied in larger quantities, the model fails to answer most questions, getting stuck in loops and replying incoherently, denoting a sharp linguistic decline. Further, at this level of noise, the model often chooses literal answers that demonstrate a lack of linguistic comprehension, failing to understand the more nuanced qualities of the question. At larger noise values, the model fails to produce intelligible language, instead replying to most prompts and questions with gibberish, or failing to reply to the prompt at all.

\textbf{Linguistic Degradation}: When smaller amounts of noise are added to the model, some word definitions and underlying meanings in the sentence are lost. In the linguistic portion of the IQ test, questions require the model to demonstrate word-finding and correlative abilities on text, with an example shown in \cref{fig:llmLingComp}. When given an example relation between words, the model is tasked with selecting another pair that shares the same relationship from a group of four words. The noiseless model likens a father to a boy the same way a mother is to a girl. With the introduction of noise with the scale of $10^{-2.8}$, the model ignores the familial aspect of the first word pair, instead likening a man to a boy, rather than its father. With a large amount of noise added, the model doesn't answer, repeating the question instead, demonstrating a breakdown in textual comprehension.

\textbf{Mathematical Degradation}: Similarly, when tasked with quantitative assessments, the model demonstrates a similar impairment of performance. Similar to the linguistic assessments, the mathematical portion demands pattern recognition and simple algebraic capabilities. \cref{fig:llmNumericComp} depicts an example of a numeric pattern completion question. The standard model answers 2 of 5 questions correctly. When noise with the scale of $10^{-2.8}$ is added to the model, it only answers 1 of 5 questions correctly, a clear loss in pattern recognition capability. However, when noise with the scale $10^{-2.575}$ is added to the model, it fails to answer each of the questions, instead replying with `22', which does not fit the pattern of any of the questions.

 \textbf{Correlation with in vivo studies}: Interestingly, in \cite{banovic2018communication} and \cite{mueller2018connected}, cognitive decline diseases first result in word-finding issues, usually before any medical diagnosis, which is then followed by difficulty with sentence syntax, the production of incomprehensible speech, and the exhibition of reduced sentence complexity. These symptoms are parallel to our model's results in the English portions of our IQ examination, through the model's subpar performance on synonym-finding and pattern recognition questions. Further, as more noise is introduced to the model, it begins to produce repetitive and incoherent speech, failing to address the question, and getting stuck in loops, as humans would when victim to cognitive decline diseases.

\section{Conclusions and Discussions}
\label{sec:conclusions}

Summing up, our work introduces a novel and versatile deep learning methodology termed ``neural erosion," a concept that transforms the dynamics of neural networks' performance degradation. This concept has been exemplified within the context of sentiment analysis and IQ Test, thereby pioneering the paradigm of employing text data for modeling cognitive diseases. The core principle of ``neural erosion" involves the controlled introduction of noise into neural network neurons during/or after the training phase, thereby systematically impairing the network's performance. We also explored other approaches, such as pruning the neurons or the synapses of the model.

Perhaps the most intriguing aspect of our findings stems from the similarity between the patterns of linguistic and mathematical degradation between humans and machine learning models. Our quantitative results dictate a similarity between the rates of loss of cognitive capability, with both humans and ML models exhibiting a gradual but eventually exponential decline in performance with the onset of neurodegeneration, as visualized in \cref{fig:mcidementia}, \cref{fig:std_noise_ablation}, and \cref{fig:iq_test}. Specific to the LLM IQ experiments, the loss in word-finding capability between humans and LLMs is particularly interesting. the gradual loss in LLaMA 2's ability to accurately synonymize and correctly define words parallels the decline in cognitive reserve due to neurodegenerative illnesses in humans. Specific examples are present in \cref{sec:LLM_appendix} 


Similarly, neural reduction entails the pruning of network connections to streamline network size and efficiency, but this does not inherently guarantee a decline in performance. In fact, studies have shown that by strategically reducing neural connections, neural networks can experience enhanced speed and performance \cite{alexos2020local}. In contrast, our approach introduces a variety of methods during/or after training to actively and controllably simulate the decline observed in cognitive disorders. Furthermore, this work's efficacy and controllability distinguish it from these alternative methods, while it emulates closely the cognitive decline characteristics exhibited in disorders. This ability to effectively induce and regulate performance decay constitutes a unique contribution to the realm of deep learning methodologies.

Moving forward, we envision multiple avenues for extending and refining this research. One potential direction involves exploring alternative noise distributions, such as the Uniform or Triangular distributions, as substitutes for the Gaussian distribution. This exploration could offer insights into the robustness and adaptability of the ``neural erosion" concept across various noise sources. Additionally, a promising avenue for future research involves validating the outcomes of our experiments against data from individuals afflicted with cognitive decline disorders. Such validation efforts could lend empirical support to the correspondence between the simulated decline in neural networks and the actual cognitive degradation observed in clinical contexts.

In summation, this work establishes ``neural erosion" as a pioneering and adaptable methodology, advancing the understanding of cognitive diseases and augmenting the toolbox of deep learning techniques for modeling complex neurological phenomena. The deliberate control over neural network performance degradation stands as a hallmark of this research, potentially yielding applications in various fields beyond sentiment analysis with CNNs and IQ tests with LLMs.

\section{Acknowledgements}

This research is partially supported by the U.S. National Science Foundation (NSF) Theoretical Physics Program, Grant No.~PHY-1915005. This work was partially performed at the Aspen Center for Physics, supported by National Science Foundation grant No.~PHY-2210452. This research was partly supported by the National Science Foundation under Grant No.~NSF PHY-1748958. This document was partially prepared using the resources of the Fermi National Accelerator Laboratory (Fermilab), a U.S. Department of Energy, Office of Science, HEP User Facility. Fermilab is managed by Fermi Research Alliance, LLC (FRA), acting under Contract No. DE-AC02-07CH11359.

\begin{appendices}

\section{Experimental Setups}
\label{subsec:experimental_setup}

\hspace{\parindent} \textbf{Optimizer and Loss Function}: To conduct the training of our model, we have adopted the Adam optimizer, a widely employed optimization algorithm available within the PyTorch framework. We use Binary Cross Entropy with Logits as our loss criterion, aligning with the binary nature of our sentiment classification task.

\textbf{Model and Data Hyperparameters}: The training of the model is carried out iteratively over a span of 100 epochs. We set the learning rate (lr) to 0.001 to facilitate the optimization process. During this training phase, we make use of a 25\% subsample extracted from the original Twitter Sentiment Analysis dataset. Importantly, this subsampling strategy preserves the initial 1:1 balance between negative and positive sentiment labels, ensuring that class proportions remain consistent across the training data. The configuration of these parameters remains consistent across all experimental trials.

\textbf{Neurodegenerative Methods}: Innovatively, our methodology introduces noise into the training iteration process. This is achieved by selectively indexing and subsequently applying noise to either a singular layer or all layers of the model concurrently. The noise is sourced from a Gaussian distribution, with a mean value fixed at zero. In the context of our primary study, we embarked upon an array of experiments, each characterized by distinct standard deviations: specifically, powers of ten ranging from 0 to -4, in increments of 0.5. Meanwhile, for the ablation study, we adopted a uniform standard deviation value of 0.1.

This comprehensive range of standard deviations was strategically chosen to scrutinize the impact of noise magnitude on our model's performance, with an emphasis on how it influences the progressive nature of our simulated neurodegeneration process. Throughout the experiment, this assortment of parameters remained consistent, providing a controlled basis for comparative analysis.

\section{Sentiment Analysis Experiments}
\label{sent_analysis_appendix}

Here we explain in detail the quantitative results of the neurodegeneration experiments that we conducted with the methods that were mentioned in \cref{sec:proposed_methodology}.

\subsection{Comparing Noise in Different Layers}
\label{subsec:comparing_layers}

\begin{figure*}[h]
\hspace*{-1.85cm}    
\includegraphics[width=1.22\textwidth]{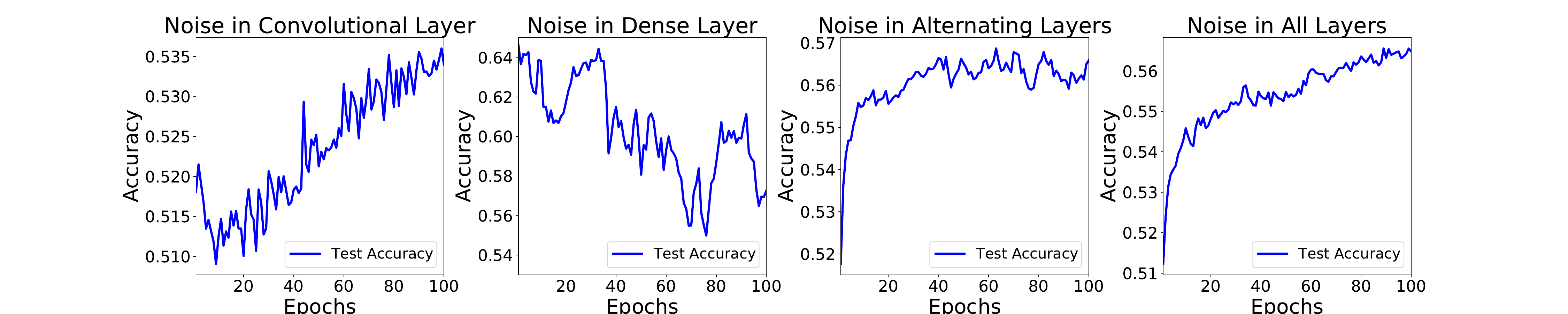}
\caption{Ablation study comparing the addition noise to different parts of the sentiment analysis model during training. All models are trained over 100 epochs, and model accuracies for the different tests from this ablation study are observable through the sub-figures.}
\label{fig:ablation_noise_layers}
\end{figure*}

\textbf{Experimental results} depicted in \cref{fig:ablation_noise_layers} delineates intriguing insights. Specifically, introducing Gaussian noise solely to the model's convolutional layers resulted in a significant degradation in performance, with an approximate accuracy of 50.4\%. Moving forward, when noise was incorporated across all layers, the observed accuracy improved to approximately 58.2\%. Meanwhile, alternating the introduction of noise across layers led to a marginal decline in accuracy, resting at around 59.9\%. Remarkably, the least substantial impact on accuracy was noted when noise was selectively added to the dense layer, registering an accuracy of approximately 68.5\%.\newline

\textbf{Validation labels} were analyzed, elucidating differences between each ablation study method and the normally trained model. While accuracy remains a pivotal benchmark, we acknowledge the significance of scrutinizing the specific labels assigned to predictions, especially in binary classification scenarios. Predicted labels between the normally trained model and the predictions generated by each noise-introduced experiment were compared in this study. Specifically, noise-injected convolutional layers yielded label discrepancies of approximately 47.1\%, while similar variations were noted in scenarios where noise was added alternately (65.5\%) or globally (66\%). The maximal divergence in predictions occurred when noise was confined to the dense layer, exhibiting label discrepancies of roughly 71.3\%. These findings accentuate the diverse impacts engendered by noise addition to various layers during model training, highlighting the nuanced interplay between noise and model prediction.

\begin{figure}
\centering
\includegraphics[width=1\textwidth]{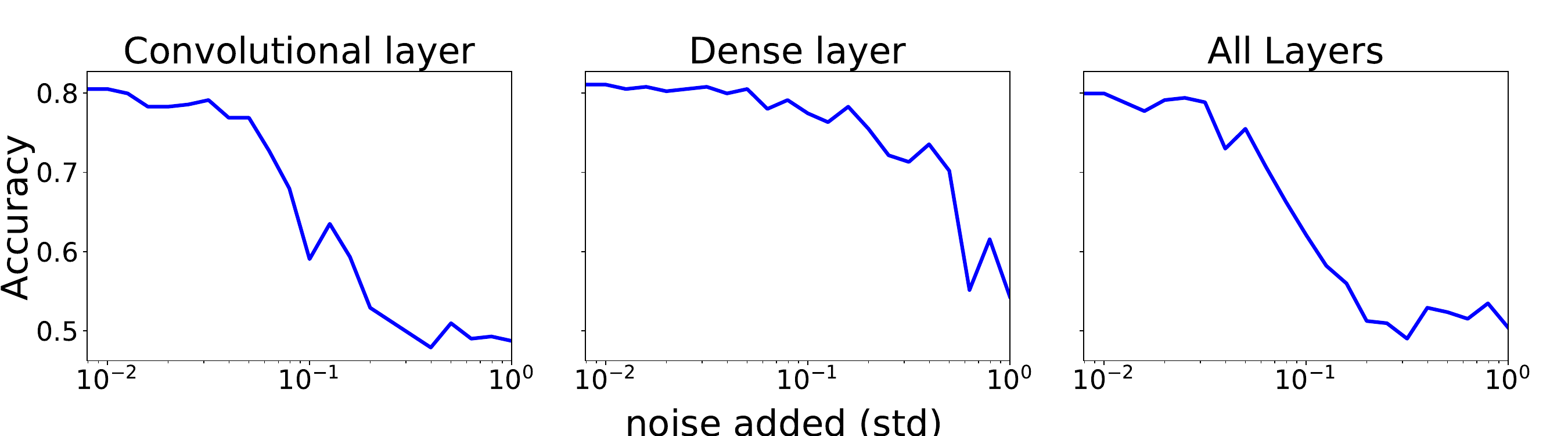}
\caption{Ablation study of adjusting the scale of the Gaussian distribution of the noise, which is only added to specific layers in a model. We find slower rates of decline when noise is added to the dense layer than to the convolutions or to all layers}
\label{fig:std_noise_after_ablation}
\end{figure}

\subsection{Analyzing post-synaptic noise}

We simulated post-synaptic noise by adding a matrix of Gaussian random noise to specific layers of the model after training was complete. We tested this by adding noise to different layers and then checked the model's validation accuracy. Our findings showed that the model with noise in the convolutional layer had the strongest correlation with the decline in accuracy over the amount of noise added, as did the model where noise was added to all layers. Interestingly, the model that had noise added to its dense layer seemed to show a slower loss of accuracy when compared to the other experiments. These findings are detailed below in \cref{fig:std_noise_after_ablation}.

\section{Combination Experiments}
\label{subsec:combinationExperiments}

\subsection{Post Synaptic Noise with Pruning and Deactivation Methods}

In building upon our previous experiments, we systematically evaluate combinations of our earlier methodologies to ascertain their cumulative impact on our model's precision. By comparing either neuronal deactivation or synaptic pruning, paired with noise that is added either during or after training, we gain a more nuanced comprehension of the trajectory and magnitude of these compounded perturbations on model accuracy. This enhanced understanding facilitates a more robust analogy between our computational model and human cognitive processes.

\subsection{Noise During Training with Pruning and Deactivation}
\label{subsec:initial_combined_exp}

\begin{figure}[h]
    \centering
    \begin{minipage}{0.8\textwidth}
        \centering
        \includegraphics[width=\textwidth,trim={4.5cm 3cm 6cm 4cm},clip]{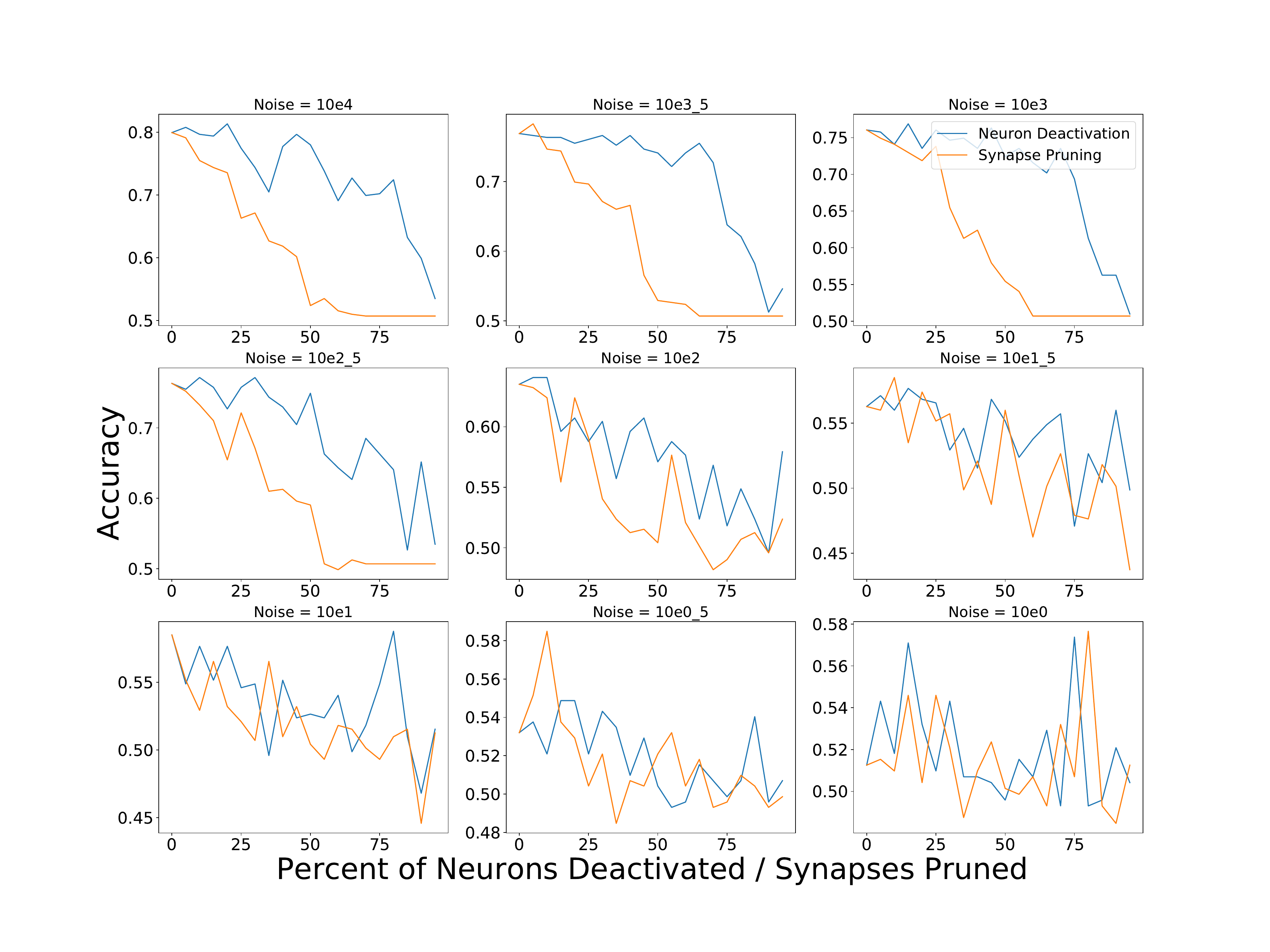}
    \end{minipage}\hfill
    \begin{minipage}{0.8\textwidth}
        \centering
        \includegraphics[width=\textwidth,trim={4.5cm 3cm 6cm 4cm},clip]{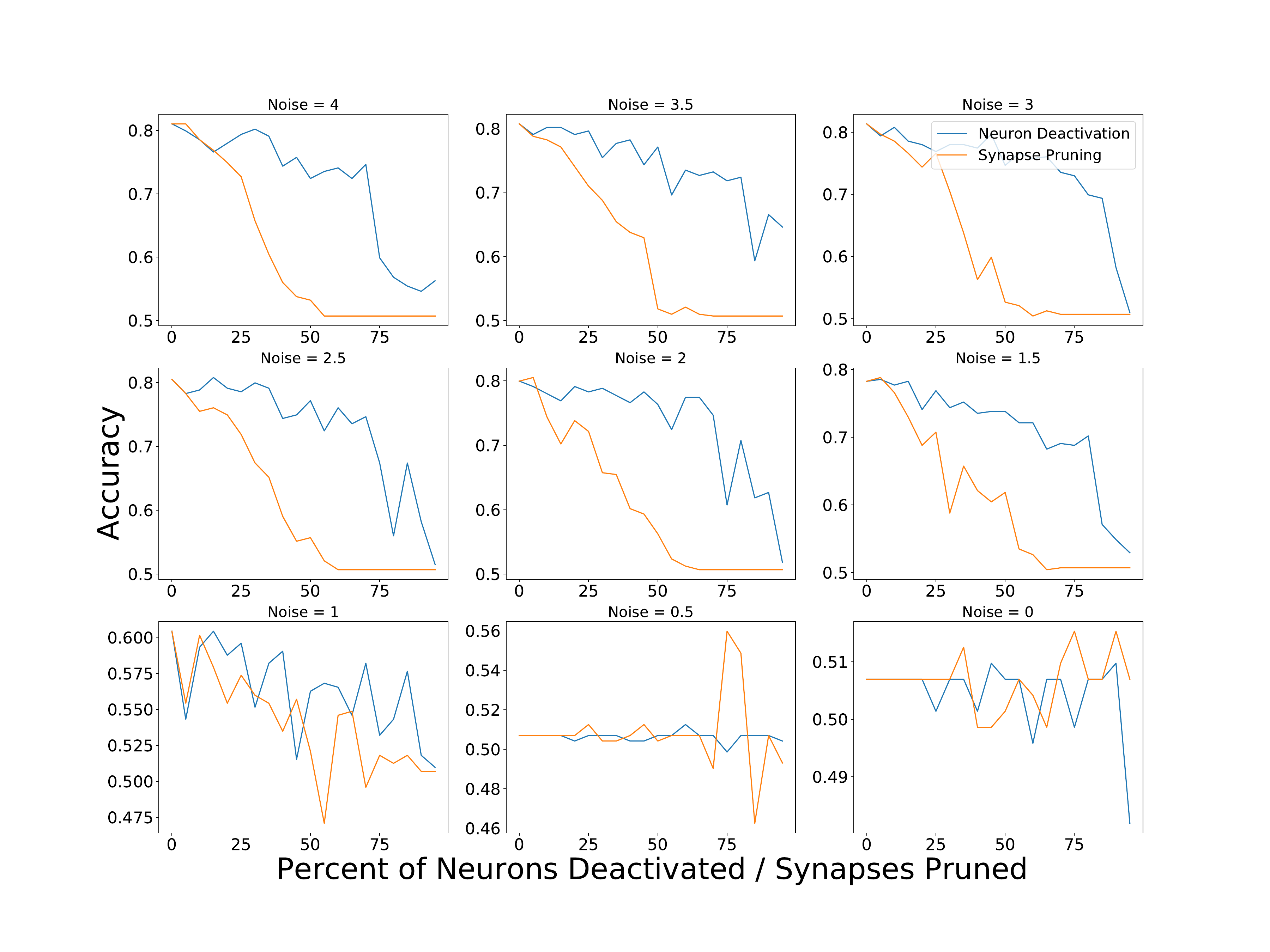}
    \end{minipage}
    \vspace{5mm}
    \caption{Ablation study comparing the effect of neuronal deactivation with synapse pruning with different methods of noise introduction, one with noise added during training (the upper nine sub-panels), and the other with noise added after training (the lower nine sub-panels).}
    \label{fig:comparingPruningAndDeactivationWithNoise}
\end{figure}

In our initial combined experimental analysis, we focus on juxtaposing the effects of neuronal deactivation and synaptic pruning in conjunction with post-synaptic noise. Notably, the model's accuracy exhibits an almost stochastic trajectory when aligned with neuronal deactivation, a stark contrast to the consistent decrement in performance observed when post-synaptic noise is paired with synaptic pruning. This differential behavior is elucidated in \cref{fig:comparingPruningAndDeactivationWithNoise}.

Furthermore, we assess the ramifications of introducing noise during the model's training phase, juxtaposing the outcomes of neuronal deactivation and synaptic pruning. Analogous to the prior experiment in \cref{subsec:initial_combined_exp}, neuronal deactivation does not manifest immediate detrimental consequences. It is only after surpassing a specific threshold that the model undergoes a marked degradation in performance. In contrast, synaptic pruning consistently exhibits a linear decrement in performance corresponding to an escalating proportion of pruned neurons. This is visualized in \cref{fig:comparingPruningAndDeactivationWithNoise}.

\clearpage

\section{LLM IQ Experiments}
\label{sec:LLM_appendix} 

\begin{figure*}[h]
    \scriptsize
    \centering
    \begin{minipage}{1\textwidth}
        \centering
        \fbox{
        \parbox{\linewidth}{
    \ttfamily 
System: This question is part of an IQ test, answer it to the best of your ability.\newline
User: Solve for the word in the brackets (first and last letter given; missing letters indicated with
                                            underline) that means the same in one sense as the word on the left and in another sense the
                                            same as the word on the right. Please keep your answers concise.\newline
VB 26: register(L \_ \_ T)lean

VB 27: obligate(T \_ \_)link

VB 28: contest(M \_ \_ \_ H)equal

VB 29: blockage ( J \_ \_ ) preserve

VB 30: whip(L \_ \_ H)tie\newline
        }}
    \end{minipage}
    \begin{minipage}[t][2in]{0.49\textwidth}
        \centering
        \fbox{\parbox{\linewidth}{
    \ttfamily 
        Assistant:  Sure, I'll do my best to answer these questions. Here are my responses:

VB 26: register(L \_ \_ T)lean
Answer: Log

VB 27: obligate(T \_ \_)link
Answer: Tie

VB 28: contest(M \_ \_ \_ H)equal
Answer: Match

VB 29: blockage ( J \_ \_ ) preserve
Answer: Jam

VB 30: whip(L \_ \_ H)tie
Answer: Lash\\\\\\\\\\
        }}
    \end{minipage}
    \hfill
    \begin{minipage}[t][2in]{0.49\textwidth}
        \centering
        \fbox{\parbox{\linewidth}{
    \ttfamily 
    Assistant:  Sure, I'll do my best to answer this question to the best of my ability.

The word in the brackets is "L\_T".

The word on the left is "L".

The word on the right is "T".

The word in the middle is "H".

The word that means the same in one sense as the word on the left is "L".

The word that means the same in another sense as the word on the right is "T".

Please let me know if you have any questions or need further clarification.
        }}
    \end{minipage}
    \caption{Above is the model's response to the English prompt when noise is added with the scale $10^{-2.575}$, and with no post-synaptic noise added. We find that the noisy model fails to understand the question, instead treating each letter like a word, and filling in the missing letter; however, it provides no reasoning for its selection besides that the letter means the same in some sense to each other letter.}

\label{fig:linguisticComparingMediumToNoNoise}
\end{figure*}

Compared to sentiment analysis models, LLMs offer more tangible textual results when subject to neurodegeneration. While quantitative results dictate that LLaMA 2 experiences cognitive decline at around the same pace as the sentiment analysis model, we are far more interested in the text-centric responses from the neurally degraded LLaMA 2 model through the IQ test experiments.

In searching for a means to control neurodegeneration in LLaMA 2, we found that adding noise only to the first half of the model's transformer blocks yields the best results. Adding noise to all transformer blocks resulted in the model completely failing with minimal noise. We reason that this is due to the lower layers being closer to the output, and therefore, even a slight perturbance would strongly affect the model's responses. Adding noise to the first half of the layers affects the model's `higher-level' understanding, and leaving the rest of the layers unmodified forces the model to reason with this corrupted internal representation and generate an input. Further, we do not conduct any experiments where noise is added during training due to the pre-trained nature of the LLaMA 2 LLM; thus, all of our experiments involve noise added after training. 

Specific to the experiments we conducted on the LLaMA 2 LLM, there was a sensitive range of $10^{-3}$ to $10^{-2.5}$ where noise added to the model had the most significant impact. Larger amounts of noise resulted in the model responding with single characters and indecipherable gibberish, while lesser amounts had no observable effect on the responses.

In smaller amounts, noise affects the textual comprehension ability of the model, as shown in the middle boxes of \cref{fig:llmLingComp} and \cref{fig:llmNumericComp}. Large amounts of noise result in the model failing to understand the prompt. In \cref{fig:linguisticComparingMediumToNoNoise}, not only does the model incorrectly answer the question, it formats its response as an indication of what letter fits in the blank, rather than the word that the blanks comprise.

\begin{figure*}
\fbox{%
  \scriptsize
    \parbox{\dimexpr\textwidth-2\fboxsep-2\fboxrule}{%
    \ttfamily 
System: This question is part of an IQ test, answer it to the best of your ability.\newline

User: You are given two terms from a four term analogy. Complete the
                                            analogy using two of the four terms in the parenthees. Only answer
                                            with the two terms.\newline
                                            VA 1: sitter is to chair as (teacup, saucer, plate, leg)\newline
                                            VA 2: needle is to thread as (cotton, sew, leader, follower)\newline
                                            VA 3: better is to worse as (rejoice, choice, bad, mourn)\newline
                                            VA 4: floor is to support as (window, glass, view, brick)\newline
                                            VA 5: veil is to curtain as (eyes, see, window, hear)\newline

> Assistant:  Sure, I'll do my best to complete the analogy using two of the four terms:\newline
VA 1: sitter is to chair as (teacup, saucer, plate)\newline
VA 2: needle is to thread as (cotton, sew, leader)\newline
The analogy is:\newline
VA 1: sitter (teacup) is to chair (seat) as VA 2: needle (cotton) is to thread.\newline

Explanation:\newline
The given terms are:\newline
* sitter (teacup)\newline
* needle (cotton)\newline

The missing terms are:\newline
* plate (leg)\newline
* leader (sew)\newline

The completed analogy would be:\newline
VA 1: sitter (teacup) is to chair (seat) as VA 2: needle (cotton) is to thread.\newline
The pattern of the analogy is:\newline
VA 1: X (teacup) is to Y (seat) as VA 2: Y (cotton) is to Z (sew)\newline
Where X, Y, and Z are the missing terms
    }%
}
  
  \caption{With large amounts of noise added to the model, we observe its linguistic capabilities break down. In this example, with noise = $10^{-2.5}$, the model selects incorrect answers and then attempts to reason with its solution by formatting the question like an equation, without addressing the rest of the questions.}
  \label{fig:linguisticHighNoise}
\end{figure*}

LLMs exhibit an entirely different set of `symptoms' when very large amounts of noise are added. \cref{fig:linguisticHighNoise} contains the LLaMA 2 LLM's responses to a linguistic question when noise with the scale of $10^{-2.5}$ is added to it. Immediately, we notice the model's inability to differentiate between different questions, combining answers from `VA 1' with `VA 2'. The model also fails to answer every question from the set, only answering and explaining question 1. Further confusion is observed from the model's questionable reasoning, stating that the question formatted with, `X is to Y, as Y is to Z', while the question is formatted as, `A is to B, as C is to D'.

\setlength{\parskip}{6pt}

\end{appendices}

\clearpage
\newgeometry{top=30mm, bottom=30mm, left=30mm, right=30mm}

\bibliographystyle{sn-nature.bst}
\bibliography{bibliography}

\end{document}